\documentclass{article}
\usepackage{spconf,amsmath,graphicx,hyperref,amsfonts}
\usepackage{booktabs}
\usepackage{pgfplots}
\pgfplotsset{compat=1.18}
\usepackage{graphicx}
\usepackage{mathtools}
\usepackage{enumerate}
\usepackage{enumitem}
\usepackage{siunitx}
\usepackage{tabularx}
\usepackage{array}
\usepackage{makecell}
\usepackage{etoolbox}

\makeatletter
\patchcmd{\@maketitle}
  {\par \vskip 1.5em}   
  {\par \vskip 0.5em}   
  {}{\typeout{Patch failed}}
\makeatother

\newcommand{\best}[1]{\textbf{#1}}
\newcommand{\second}[1]{\underline{#1}}
\newcolumntype{A}{S[table-format=2.1]} 
\newcolumntype{N}{S[table-format=3.1]} 

\definecolor{DifficultyMain}{HTML}{0072B2}
\definecolor{DifficultyCompressed}{HTML}{D55E00}
\definecolor{DifficultyRaw}{HTML}{666666}

\pgfplotsset{
  difficulty accuracy axis/.style={
    width=.31\columnwidth,
    height=.36\columnwidth,
    scale only axis,
    symbolic x coords={L,M,H},
    xtick=data,
    xticklabels={Easy,Medium,Hard},
    xticklabel style={font=\scriptsize},
    xlabel={Difficulty},
    xlabel style={font=\scriptsize,yshift=1mm},
    enlarge x limits=.22,
    ymin=40,
    ymax=82,
    ytick={40,50,60,70,80},
    ylabel={Accuracy (\%)},
    ylabel style={font=\scriptsize},
    tick label style={font=\scriptsize},
    axis lines=box,
    xtick pos=bottom,
    ytick pos=left,
    axis y discontinuity=none,
    axis line style={black!65,line width=.55pt},
    tick style={black!65,line width=.45pt},
    tick align=outside,
    ymajorgrids,
    grid style={black!14,line width=.4pt},
    clip=false,
  },
  difficulty tokens axis/.style args={#1}{
    at={(#1.south west)},
    anchor=south west,
    width=.31\columnwidth,
    height=.36\columnwidth,
    scale only axis,
    symbolic x coords={L,M,H},
    enlarge x limits=.22,
    axis x line=none,
    xtick=\empty,
    axis y line*=right,
    ytick pos=right,
    axis line style={black!65,line width=.55pt},
    tick style={black!65,line width=.45pt},
    tick align=outside,
    ymin=150,
    ymax=620,
    ytick={200,300,400,500,600},
    ylabel={Mean output tokens},
    ylabel style={font=\scriptsize},
    tick label style={font=\scriptsize},
    clip=false,
  },
  main bars/.style={
    ybar,bar width=3.8pt,bar shift=-4.4pt,
    draw=DifficultyMain,fill=DifficultyMain!72,line width=.55pt,
  },
  compressed bars/.style={
    ybar,bar width=3.8pt,bar shift=0pt,
    draw=DifficultyCompressed,fill=DifficultyCompressed!58,line width=.55pt,
  },
  raw bars/.style={
    ybar,bar width=3.8pt,bar shift=4.4pt,
    draw=DifficultyRaw,fill=DifficultyRaw!38,line width=.55pt,
  },
  main line/.style={
    color=DifficultyMain,thick,dashed,
    mark=*,mark size=2.25pt,
    mark options={solid,fill=white,line width=.75pt},
  },
  compressed line/.style={
    color=DifficultyCompressed,thick,dashed,
    mark=square*,mark size=2.05pt,
    mark options={solid,fill=white,line width=.7pt},
  },
  raw line/.style={
    color=DifficultyRaw,thick,thick,dashed,
    mark=triangle*,mark size=2.35pt,
    mark options={solid,fill=white,line width=.7pt},
  },
}

\newif\ifshow
\showtrue

\newcommand{\MH}[1]{%
  \ifshow
    \textcolor{blue}{(MH: #1)}%
  \fi
}

\showfalse

\title{SWEAR: Section-Wise Efficient Audio Reasoning via Self-Rated Difficulty}

\name{\textbf{Yixuan Li}$^{1, 2}$ \quad \textbf{Minghao Lv}$^{1}$ \quad \textbf{Xuenan Xu}$^{3}$ \quad \textbf{Zihao Zheng}$^{1}$ \quad \textbf{Mengyue Wu}$^{1}$\sthanks{
    Corresponding author.
  }}
\address{
    $^1$ MoE Key Lab of Artificial Intelligence, X-LANCE Lab, Shanghai Jiao Tong University, China \\
    $^2$ Shanghai AI Lab, China \qquad $^3$ KlingAI Research, China
}

\begin{document}

\setlength{\textfloatsep}{10pt plus 2pt minus 2pt}
\setlength{\dbltextfloatsep}{10pt plus 2pt minus 2pt}
\setlength{\intextsep}{8pt plus 2pt minus 2pt}

\setlength{\abovedisplayskip}{3.5pt}
\setlength{\belowdisplayskip}{3.5pt}


\maketitle


%
\begin{abstract}
Recent work on large audio-language models (LALMs) explores adaptive reasoning to balance accuracy and efficiency. However, existing approaches often rely on coarse-grained difficulty estimation, overlooking the diverse factors that determine reasoning effort. We address this limitation by deriving five-dimensional difficulty scores and compressed reasoning traces from complementary offline assessments. Based on this characterization, we propose SWEAR, a difficulty-aware reasoning framework that learns difficulty signals during supervised fine-tuning and uses reinforcement learning to adapt reasoning allocation across reasoning stages. Experiments on MMAU-test-mini and MMAR show that SWEAR improves accuracy while substantially reducing generated tokens, shortening reasoning for easy instances without sacrificing performance on difficult ones.
\end{abstract}
\vspace{-0.5em}

\section{Introduction}
\label{sec:intro}
\vspace{-0.5em}


LALMs increasingly use explicit reasoning to connect acoustic observations with answers~\cite{ma2025audiocot,wen2025sari,tian2025stepaudior1}.
As reasoning incurs additional computational cost, characterizing reasoning difficulty is essential for adaptive reasoning allocation, allowing more budget for challenging cases while reducing unnecessary computation for easier ones.
However, accurately estimating input difficulty and determining the corresponding reasoning effort remain challenging, motivating difficulty-aware reasoning strategies that dynamically adjust reasoning allocation.

Addressing this question, a parallel line of work has explored input-dependent reasoning effort rather than fixed CoT behaviors. In text reasoning, AdaptThink~\cite{zhang-etal-2025-adaptthink}, AutoThink~\cite{tu2025learning}, and related methods~\cite{shen-etal-2025-dast} adapt reasoning length or computational budgets based on instance difficulty. Similar ideas have been investigated in visual reasoning, where FAST~\cite{xiao2025fastslow} and CAR~\cite{lu2025prolonged} adjust reasoning effort according to input cues. These studies demonstrate the potential of adaptive reasoning for balancing accuracy and efficiency across modalities.

Recent studies have explored difficulty-aware reasoning adaptation in LALMs.
Audio-Thinker~\cite{wu2026audiothinker} and Omni-AutoThink~\cite{yang2025omniautothink} use reinforcement learning to adapt reasoning activation, while Think Smart, Not Hard~\cite{sheng-etal-2026-think} and CESAR~\cite{fan2026cesar} adjust reasoning length based on difficulty-related signals.
However, these approaches typically characterize difficulty with a single overall measure, such as rollout performance, attention statistics, or trace-level rewards.
Such global estimates indicate whether more reasoning is needed but provide limited guidance on where to allocate reasoning effort.
This limitation is important for audio reasoning, where different stages, including question interpretation, acoustic evidence extraction, and answer inference, may introduce distinct and uncorrelated difficulties.
Therefore, fine-grained and multi-dimensional difficulty characterization across reasoning sections is needed for adaptive allocation of reasoning effort within structured reasoning traces.

To address this gap, we propose SWEAR,
a framework for \textbf{S}ection-\textbf{W}ise \textbf{E}fficient \textbf{A}udio \textbf{R}easoning. In addition, We built \textsc{DiCoTA} from
CoTA~\cite{zhifei-etal-2025-audio}, which supplies section-wise supervision and
compressed traces from five offline assessments. SWEAR
learns the ratings during Supervised Fine-Tuning (SFT) and uses them during Reinforcement Learning (RL) to control section-wise
reasoning length and reference-policy regularization, together with a
feedback-driven curriculum. This encourages concise reasoning on easy inputs
while preserving sufficient flexibility for challenging ones.

\begin{figure*}[t]
  \centering
  \includegraphics[width=\textwidth]{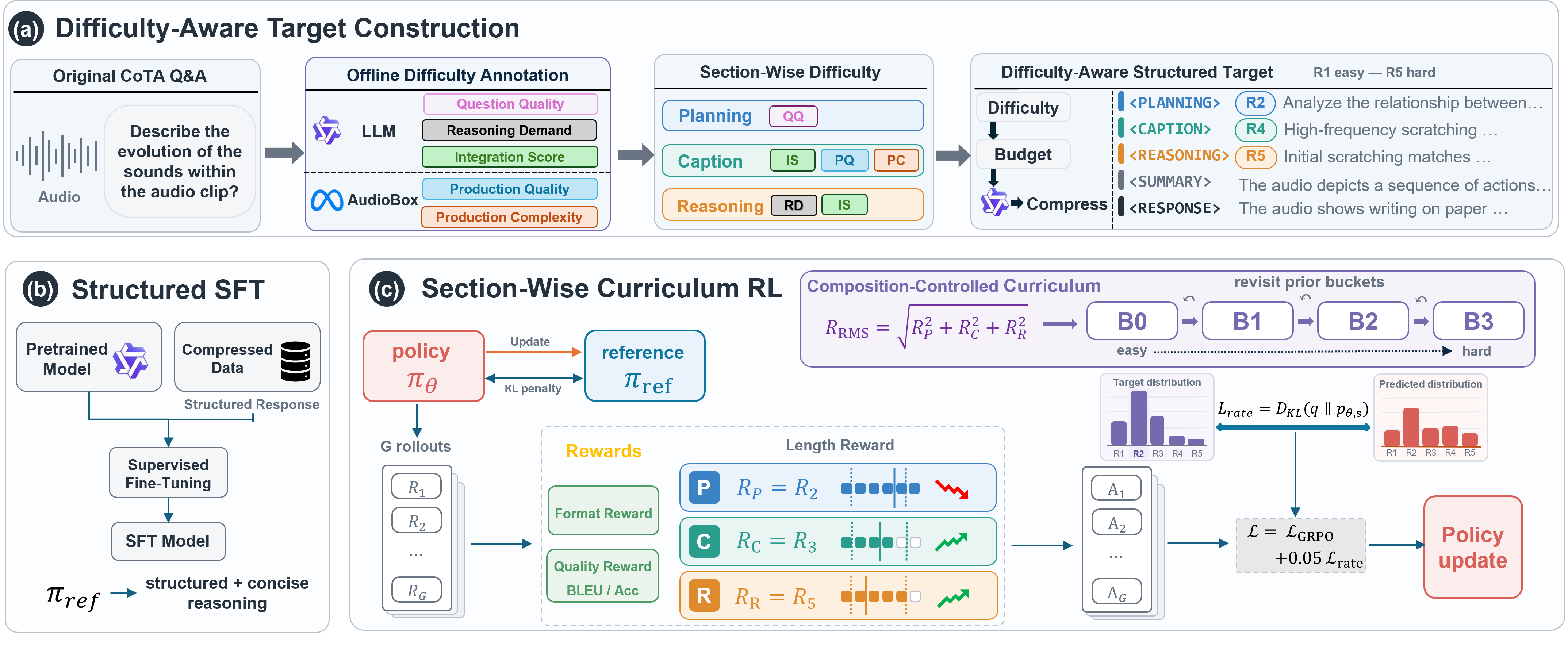}
  \vspace*{-2em}
  \caption{Overview of our multi-dimensional difficulty-aware audio reasoning framework. (a) Offline linguistic and acoustic annotations estimate Planning, Caption, and Reasoning difficulty, which determine section ratings and compression budgets for structured targets. (b) Structured SFT teaches the model to produce concise, rated reasoning traces. (c) Curriculum RL progresses from easy to hard while jointly optimizing task quality, format, rating-conditioned section-length rewards, and teacher-anchored rating calibration under reference-policy KL regularization. Offline annotators are used only during training.
}
  \label{fig:overview}
\end{figure*}

Overall, our contributions are summarized as follows:
\begin{itemize}[
  leftmargin=1em,
  topsep=2pt,
  partopsep=0pt,
  parsep=0pt,
  itemsep=0.2pt
]
\item We construct \textsc{DiCoTA}, a 575K-example audio reasoning dataset with five-dimensional difficulty scores conditioned on reasoning sections and compressed reasoning traces.

\item We develop SWEAR, a difficulty-aware curriculum learning framework for multi-dimensional reasoning adaptation. SWEAR learns difficulty self-ratings during SFT and leverages them in RL for adaptive reasoning length control.

\item Experiments on MMAU and MMAR show that SWEAR improves accuracy by 2.6 and 6.3 percentage points while reducing generated tokens by 60.1\% and 62.3\%, respectively, compared with SFT without difficulty supervision.



\end{itemize}

\vspace{-1em}

\section{Method}
\label{sec:method}

\vspace{-0.5em}

\textsc{SWEAR} learns to allocate reasoning effort across
trace sections through a three-stage pipeline, as illustrated in
Fig.~\ref{fig:overview}.
First, we construct \textsc{DiCoTA}, an audio reasoning
dataset with multi-dimensional difficulty annotations, by
assigning difficulty scores to individual sections and
compressing their content under difficulty-dependent budgets.
\MH{what is the difference between "function" and "section".} and compressing each section under an adaptive budget. 
Second, SFT teaches the policy to predict each section's
difficulty rating before generating its content. Finally, difficulty-conditioned
curriculum RL progressively aligns section-wise reasoning effort with the
policy's own difficulty predictions.

\vspace{-1em}

\subsection{DiCoTA: Difficulty-Aware Trace Construction}
\label{sec:trace_construction}

We construct \textsc{DiCoTA}, a difficulty-annotated extension
of CoTA~\cite{zhifei-etal-2025-audio}. Each trace comprises
Planning ($P$), Caption ($C$), Reasoning ($R$), and Summary ($S$),
followed by a Response ($A$). We assign difficulty targets to
$P$, $C$, and $R$, while $S$ uses a fixed consolidation budget.


We characterize audio reasoning difficulty through five dimensions spanning acoustic quality, audio complexity, and reasoning demand: production quality (PQ), production complexity (PC), integration score (IS), reasoning demand (RD), and question quality (QQ). PQ and PC are obtained from Audiobox Aesthetics~\cite{tjandra2025audiobox}, measuring technical quality and the complexity of audio components, respectively. IS, RD, and QQ are provided by Qwen3-Omni-30B-A3B~\cite{xu2025qwen3omni}, capturing reliance on fine-grained acoustic evidence, inference complexity, and question formulation quality. Together, these dimensions characterize complementary sources of difficulty, ranging from acoustic perception and evidence extraction to question understanding and answer reasoning.

With $g_m(x)=(x-1)/(m-1)$ for an $m$-point score, we construct
difficulty proxies by reversing QQ and PQ: lower question clarity
and audio quality receive higher proxy scores.
\begin{equation}
\begin{gathered}
q = 1-g_5(\mathrm{QQ}), \qquad
i = g_5(\mathrm{IS}), \qquad
r = g_5(\mathrm{RD}), \\
p = 1-g_{10}(\mathrm{PQ}), \qquad
c = g_{10}(\mathrm{PC}).
\end{gathered}
\label{eq:aligned_scores}
\end{equation}

We define the integrated difficulty of each trace section by
aggregating the relevant normalized proxies according to its
functional role:
\begin{equation}
D_P=q,\qquad
D_C=\tfrac13(i+p+c),\qquad
D_R=\tfrac12(r+i).
\label{eq:difficulty}
\end{equation}

Specifically, $D_P$ captures question interpretation difficulty. $D_C$ reflects acoustic description difficulty by considering task-relevant evidence, audio quality, and scene complexity. $D_R$ characterizes reasoning difficulty through inference demand and acoustic evidence requirements.

\MH{Did we mention that $D$ is the integrated difficulty of each section?}

\MH{should we put this paragraph to $D_S$ part? We first introduce continus $D_S$,then quantized $b_s$ and the data format using $b_s$. Finally we use $B_s$ to process dataset. Its a bit confused for me that we put processing dataset para in the middle. We can say "In order to be more accurate, we use continus $D_s$ not quantized $b_s$". btw, did we cite fig 1 in the whole paper? We should cite even a single sentence.}
The continuous difficulties are quantized into five equal-width levels, where $b_s$ denotes the corresponding discrete difficulty rating index for section $s\in\mathcal{G}=\{P,C,R\}$.
\begin{equation}
b_s=\min{\{5,\lfloor5D_s\rfloor+1\}}
\end{equation}
represented by rating tokens
$\mathsf{R}_1,\ldots,\mathsf{R}_5$. The resulting target response is
\begin{equation}
y=
\langle P\rangle\langle\mathsf{R}_{b_P}\rangle y_P;
\langle C\rangle\langle\mathsf{R}_{b_C}\rangle y_C;
\langle R\rangle\langle\mathsf{R}_{b_R}\rangle y_R;
\langle S\rangle y_S;
\langle A\rangle y_A.
\label{eq:structured_output}
\end{equation}

Unlike uniform trace compression, we allocate a fixed trace budget according to the difficulty of different sections. Let $B$ denote the total budget for $P$, $C$, $R$, and $S$ (excluding the final answer), and let $B_S$ be the fixed Summary budget. For $s\in\mathcal{G}$, the target budget is
\begin{equation}
B_s=(B-B_S)
\frac{D_s+\epsilon}
{\sum_{j\in\mathcal{G}}(D_j+\epsilon)},
\qquad \epsilon>0,
\label{eq:teacher_budget}
\end{equation}
where the difficulty floor prevents degenerate zero allocations. We use continuous difficulty scores $D_s$ rather than quantized ratings $b_s$ for finer-grained budget allocation. Conditioned on $(D_s,B_s)$, the teacher removes redundant steps while preserving each section's function and consistency with the answer. This allocation assigns more tokens to difficult sections while producing more concise traces for easier ones.

Given an audio--question input $x$ and its structured target $y$, we perform
SFT to learn the target response generation. Since
each rating is predicted before its corresponding section, it serves as a
causal conditioning token rather than a post-hoc annotation. SFT thus
establishes the section-wise difficulty interface later optimized by
curriculum RL.


\MH{it is occur to me that we should add the statisic of DiCoTA.}
\vspace{-1em}
\subsection{Difficulty-Adaptive Curriculum RL}
\label{sec:section_grpo}

To encourage difficulty-aware reasoning adaptation and improve the efficiency of reasoning generation, we optimize the SFT policy with the clipped Group Relative Policy Optimization (GRPO) objective~\cite{shao2024deepseekmath}, using group-normalized advantages and KL regularization to the frozen SFT reference.
\begin{equation}
\mathcal{L}_{\mathrm{GRPO}}
=
\mathcal{L}_{\mathrm{clip}}
-\beta_iD_{\mathrm{KL}}
(\pi_\theta\|\pi_{\mathrm{SFT}}),
\end{equation}
where \(\mathcal{L}_{\mathrm{clip}}\) denotes the clipped policy objective with group-normalized advantages.

\noindent\textbf{Reward Function.}
For a sampled response $y_i$, we define
\begin{equation}
r_i= r_{\mathrm{fmt}}(y_i)
+ r_{\mathrm{qual}}(y_i)
+ r_{\mathrm{len}}(y_i),
\label{eq:reward}
\end{equation}
where $r_{\mathrm{fmt}}$ checks the required section and rating-token format,
and $r_{\mathrm{qual}}$ uses normalized BLEU for speech translation or answer
accuracy for closed-choice tasks.

For each section $s\in\mathcal{G}$, let $L_{i,s}$ and $C_{i,s}$ denote
the generated and compressed \textsc{DiCoTA} reference lengths,
respectively. Given the predicted rating $b_{i,s}\in\{1,\ldots,5\}$,
we define a rating-dependent target length ratio $\tau_{i,s}$, which
increases with $b_{i,s}$, and compute the length reward as
\begin{equation}
 r_{\mathrm{len}}(y_i)
=-\sum_{s\in\mathcal{G}}
\left(\frac{L_{i,s}}{C_{i,s}}-\tau_{i,s}\right)^2.
\label{eq:length_reward}
\end{equation}
Here $\tau_{i,s}$ is the target ratio between the generated and
compressed reference lengths for section $s$; higher ratings favor
longer generated sections.


\noindent\textbf{Feedback-driven curriculum.}
We use task-quality feedback to gradually introduce harder samples.
The samples are ranked by
$D_{\mathrm{rms}}=\sqrt{(D_P^2+D_C^2+D_R^2)/3}$ and divided into $K_c$
equal-frequency buckets from easy to hard. The competence score $c\in[0,1]$ is a scalar estimate
of the model's mastery of the current bucket, updated from recent reward
feedback.
Following AdaCuRL~\cite{li2026adacurl}, training starts with $\mathcal{B}_1$
and a competence score $c^{(0)}=0$\MH{did we mention competence score before? We can't take it for granted that reviewer will read Adacurl. We can simply define competence score like "competence score $c$ is defined as a representative of model ability in a certain window $M$" or more accurate defination. Then we can say "competence score $c$ start with $c^{(0)}=0$ and updated as ...".}, updated as
\begin{equation}
\resizebox{0.98\columnwidth}{!}{$
c^{(t+1)}
=\operatorname{clip}_{[0,1]}\!\left(
c^{(t)}
+\bigl(\bar r_{\mathrm{qual}}^{(t)}-0.5\bigr)
\max\{1-c^{(t)},\gamma\}
\right).
$}
\label{eq:competence}
\end{equation}
Here, $\bar r_{\mathrm{qual}}^{(t)}$ averages the
latest $M$ prompt-level quality scores from the
hardest active bucket, each computed by averaging
the quality rewards of that prompt's rollouts, and $\gamma$ sets a minimum
update scale. When $c\geq(k-1)/K_c$\MH{and what is $K_c$}, we activate the next bucket
$\mathcal{B}_k$ and reset $c$ to that threshold. Earlier buckets remain
active with exponentially downweighted sampling.


\noindent\textbf{Adaptive KL regularization.}
We adapt the KL coefficient as $\beta_i=\beta_0(1-0.5R_{\mathrm{RMS}})$,
where $R_{\mathrm{RMS}}$ aggregates the normalized
\textit{predicted} section ratings. Higher predicted difficulty allows
greater deviation from the frozen SFT reference.


\noindent\textbf{Rating calibration.}
To improve the accuracy of predicted difficulty, we calibrate the control
ratings against the unquantized offline difficulty. Let $\mu_s=5D_s\in[0,5]$ (without
rounding) be the unquantized offline difficulty, and define
$q_s(j)\propto\exp[-(j-\mu_s)^2/(2\sigma^2)]$
over the integer rating levels $j\in\{1,\ldots,5\}$, normalized across these
levels.
We obtain $p_{\theta,s}$ by applying softmax only to
the logits of the rating
tokens $1$--$5$ immediately before the
rating digit, conditioned on the audio, question,
and preceding generated content.
The calibration loss for section $s$ is
$\mathcal{L}_{\mathrm{cal},s}
=D_{\mathrm{KL}}(q_s\Vert p_{\theta,s})$.


The final loss is
$\mathcal{L}=\mathcal{L}_{\mathrm{GRPO}}
+0.05\sum_{s\in\mathcal{G}}\mathcal{L}_{\mathrm{cal},s}$.
At inference, the model generates ratings, sections,
and the answer autoregressively, without external
difficulty scoring or reference traces.

\vspace{-1em}

\begin{table*}[ht]
\centering
\small
\setlength{\tabcolsep}{2.5pt}
\renewcommand{\arraystretch}{1}

\resizebox{\textwidth}{!}{%
\begin{tabular}{
  @{}l
  *{8}{c}
  @{\hspace{8pt}}
  *{8}{c}
  @{}
}
\toprule
& \multicolumn{8}{c}{MMAU-test-mini}
& \multicolumn{8}{c}{MMAR} \\
\cmidrule(lr){2-9}
\cmidrule(lr){10-17}

Model
& Easy & Medium & Hard
& Sound & Music & Speech
& Overall & Token
& Easy & Medium & Hard
& Sound & Music & Speech
& Overall & Token \\
\midrule

Qwen2.5-Omni-7B~\cite{xu2025qwen25omni}
& 73.87 & 70.96 & 61.26
& 71.47 & 66.47 & 68.17
& 68.7 & 14.34
& 60.18 & 60.36 & 56.16
& 58.18 & 50.00 & 62.93
& 58.9 & 13.18 \\

\midrule
\multicolumn{17}{@{}l}{\textit{Public Baselines}} \\
\addlinespace[1pt]

Audio-Reasoner-7B~\cite{zhifei-etal-2025-audio}
& 69.37 & 69.16 & 61.26
& 66.67 & 64.67 & 68.77
& 66.7 & 504.36
& 49.40 & 43.84 & 43.84
& 45.45 & 39.81 & 48.98
& 45.6 & 537.18 \\

Ke-Omni-R-7B~\cite{zhao2025keomnir}
& \best{79.88} & 75.75 & 64.86
& 78.08 & 70.36 & 72.07
& \best{73.5} & \best{76.64}
& 58.38 & 56.46 & 49.85
& 55.15 & 42.23 & 58.84
& 55.0 & \best{70.34} \\

\makecell[l]{Audio-Flamingo 3~\cite{ghosh2025audioflamingo3}}
& \second{78.68} & \best{76.65} & \second{65.17}
& 77.48 & 69.46 & 73.27
& \second{73.4} & 199.90
& 61.68 & 55.26 & 45.95
& 49.70 & 48.06 & 58.16
& 54.5 & \second{189.10} \\

Omni-CLST~\cite{zhao2025omniclst}
& 77.78 & 74.85 & 63.66
& 77.48 & \second{71.56} & 67.27
& 72.1 & 204.03
& 60.78 & 60.06 & \second{56.46}
& \best{63.49} & 52.45 & 56.17
& 59.1 & 368.01 \\

Omni-R1~\cite{rouditchenko2025omnir1}
& 77.18 & 73.65 & \second{68.47}
& \best{80.48} & 67.96 & 70.87
& 73.2 & 7.11
& \best{66.17} & \second{63.66} & 54.37
& 58.81 & \second{53.46} & \best{65.14}
& \second{61.4} & 6.82 \\

\midrule
\multicolumn{17}{@{}l}{\textit{Our Models}} \\
\addlinespace[1pt]

CoTA-SFT
& 75.68 & 74.25 & 62.46
& 71.17 & 67.37 & \second{74.17}
& 70.9 & 494.58
& 59.28 & 57.06 & 50.76
& 56.36 & 45.63 & 59.86
& 55.7 & 519.82 \\

DiCoTA-SFT
& 74.17 & \second{76.35} & 65.47
& 70.27 & 69.76 & \best{75.38}
& 71.8 & 216.18
& 58.68 & 56.46 & 53.45
& 56.36 & 48.06 & 57.82
& 56.0 & 222.74 \\

SWEAR
& \second{78.68} & 72.46 & \best{69.37}
& \second{78.38} & \best{73.65} & 68.47
& \best{73.5} & \second{197.13}
& \second{62.28} & \best{66.07} & \best{57.66}
& \second{62.42} & \best{54.85} & \second{64.29}
& \best{62.0} & 195.85 \\

\bottomrule
\end{tabular}%
}
\vspace{-1em}
\caption{
Accuracy (\%) across predicted difficulty levels and audio categories on
MMAU-test-mini and MMAR.
Overall denotes accuracy on the full test set, and Token denotes the average
number of generated tokens.
The best and second-best results are highlighted in \textbf{bold} and
\underline{underlined}, respectively.
Token rankings include only models with explicit CoT.
}
\label{tab:main-predicted-difficulty}
\end{table*}

\section{Experiment and Results}
\vspace{-0.5em}
\subsection{Experimental Setup}

\noindent\textbf{Training Data.}
From \textsc{DiCoTA}, we use 402,496 examples for SFT and a separate set of
113,410 reward-compatible examples for RL. The RL set covers speech-to-text transformation and multiple-choice audio reasoning tasks, split by $D_{\mathrm{rms}}$ into four equal-sized difficulty buckets.

\noindent\textbf{Evaluation.}
We evaluate on the 1,000-example test sets of MMAU-test-mini and MMAR, using answer
accuracy as the effectiveness metric and average generated tokens as the
efficiency metric. For difficulty, we score each benchmark example using
the same procedure as \textsc{DiCoTA}, partition each benchmark
independently into 3 equal-sized groups from easy to hard. \MH{should add the number of each group. And also for difficulty buckets in the begin of 3.1?}Answers are extracted by jointly matching the predicted
option label and option content. We compare with representative open-source audio reasoning baselines and SFT variants based on CoTA and DiCoTA, denoted as CoTA-SFT and DiCoTA-SFT, respectively. All models follow the same generation and
scoring protocol.

\noindent\textbf{Training Settings.}
We use Qwen2.5-Omni-7B as the backbone. SFT is conducted for one epoch with
LoRA rank 64, scaling factor 128, a global batch size of 64, and a learning
rate of $1\times10^{-5}$. RL continues LoRA optimization from the SFT
checkpoint for 2,000 steps with a learning rate of $1\times10^{-6}$. Each
prompt samples $G=6$ responses, and the curriculum uses $K_c=4$ difficulty buckets.
For the length reward in Eq.~\eqref{eq:length_reward}, we set
$\tau_{i,s}=(b_{i,s}+27)/30$, which increases linearly from $0.93$ to $1.07$
as $b_{i,s}$ goes from $1$ to $5$.

\vspace{-1em}
\subsection{Results and Analysis}

As shown in Table~\ref{tab:main-predicted-difficulty}, our full model achieves 73.5\% accuracy on MMAU-test-mini, tying the best result, and leads on MMAR with 62.0\%. These gains are achieved with only 197.13 and 195.85 generated tokens, reducing token usage by 60.1\% and 62.3\% relative to CoTA-SFT. These results demonstrate that SWEAR achieves a stronger accuracy-efficiency trade-off than prior explicit-CoT audio reasoning models, including Audio-Reasoner, Audio-Flamingo 3, Omni-CLST, and CoTA-SFT, consistently reducing reasoning tokens while maintaining or improving task performance.

\vspace{-0.5em}
\begin{figure}[ht]
  \centering

  \begin{tikzpicture}
    \begin{axis}[
      name=mmau-accuracy,
      difficulty accuracy axis,
      at={(0pt,0pt)},
      anchor=south west,
    ]
      \addplot[main bars]
        coordinates {(L,72.46) (M,78.68) (H,69.37)};
      \addplot[compressed bars]
        coordinates {(L,76.35) (M,74.17) (H,65.47)};
      \addplot[raw bars]
        coordinates {(L,74.25) (M,75.68) (H,62.46)};
    \end{axis}

    \begin{axis}[
      difficulty tokens axis={mmau-accuracy},
      axis y line=none,
      ytick=\empty,
      ylabel={},
    ]
     \addplot[main line]
       coordinates {(L,172.81) (M,196.13) (H,222.45)};   
    
     \addplot[compressed line]
       coordinates {(L,188.99) (M,220.10) (H,239.48)};   
    
     \addplot[raw line]
       coordinates {(L,414.07) (M,513.82) (H,555.85)};   
    \end{axis}

    \begin{axis}[
      name=mmar-accuracy,
      difficulty accuracy axis,
      at={(.40\columnwidth,0pt)},
      anchor=south west,
      yticklabels=\empty,      
      ytick style={draw=none},  
      ylabel={}
    ]
      \addplot[main bars]
        coordinates {(L,62.28) (M,66.07) (H,57.66)};
      \addplot[compressed bars]
        coordinates {(L,58.68) (M,56.46) (H,53.45)};
      \addplot[raw bars]
        coordinates {(L,59.28) (M,57.06) (H,50.76)};
    \end{axis}

    \begin{axis}[
      difficulty tokens axis={mmar-accuracy},
      legend columns=1,
      legend style={
        at={(axis description cs:.99,.75)},
        anchor=north east,
        draw=black!55,
        fill=white,
        fill opacity=.92,
        text opacity=1,
        line width=.4pt,
        rounded corners=1pt,
        font=\fontsize{4}{5}\selectfont,
        inner sep=-.2pt,
        row sep=-2.5pt,
        cells={anchor=west},
        /tikz/every even column/.append style={column sep=.9pt},
      },
    ]
      \addplot[main line,forget plot]
       coordinates {(L,174.35) (M,190.01) (H,223.20)};   
    
     \addplot[compressed line,forget plot]
       coordinates {(L,209.06) (M,223.33) (H,235.83)};   
    
     \addplot[raw line,forget plot]
       coordinates {(L,469.17) (M,517.37) (H,572.92)};   

      \addlegendimage{
        legend image code/.code={
          \filldraw[
            draw=DifficultyMain,
            fill=DifficultyMain!72,
            line width=.3pt,
            rounded corners=0pt
          ] (1mm,-.45mm) rectangle (2mm,.45mm);   
          \draw[DifficultyMain,thick,dashed]
          plot[
            mark=*,
            mark size=.75pt,
            mark repeat=2,
            mark phase=2,
            mark options={
              solid,
              fill=white,
              line width=.5pt
            }
          ] coordinates {
            (1.8mm,0)
            (3.3mm,0)
            (4.8mm,0)
          };
        }
      }
      \addlegendentry{\scalebox{.9}{SWEAR}}

      \addlegendimage{
        legend image code/.code={
          \filldraw[
            draw=DifficultyCompressed,
            fill=DifficultyCompressed!58,
            line width=.3pt,
            rounded corners=0pt
          ] (1mm,-.45mm) rectangle (2mm,.45mm); 
          \draw[DifficultyCompressed,thick,dashed]
          plot[
            mark=square*,
            mark size=.8pt,
            mark repeat=2,
            mark phase=2,
            mark options={
              solid,
              fill=white,
              line width=.5pt
            }
          ] coordinates {
            (1.8mm,0)
            (3.3mm,0)
            (4.8mm,0)
          };
        }
      }
      \addlegendentry{\scalebox{.9}{DiCoTA-SFT}}

      \addlegendimage{
        legend image code/.code={
          \filldraw[
            draw=DifficultyRaw,
            fill=DifficultyRaw!38,
            line width=.3pt,
            rounded corners=0pt
          ] (1mm,-.45mm) rectangle (2mm,.45mm);
          \draw[DifficultyRaw,thick,dashed]
          plot[
            mark=triangle*,
            mark size=.9pt,
            mark repeat=2,
            mark phase=2,
            mark options={
              solid,
              fill=white,
              line width=.5pt
            }
          ] coordinates {
            (1.8mm,0)
            (3.3mm,0)
            (4.8mm,0)
          };
        }
      }
      \addlegendentry{\scalebox{.9}{CoTA-SFT}}
    \end{axis}

    \path (mmau-accuracy.north) -- (mmar-accuracy.north)
      node[midway, above=6mm, font=\footnotesize, text=black!80]
      {\textbf{Bars:} Accuracy (\%) \quad \textbf{Lines:} Mean output tokens};
    
    \node[anchor=south,font=\small\bfseries,yshift=1mm]
      at (mmau-accuracy.north) {(a) MMAU};
    \node[anchor=south,font=\small\bfseries,yshift=1mm]
      at (mmar-accuracy.north) {(b) MMAR};
  \end{tikzpicture}
  \vspace{-2.5em}
  \caption{Accuracy (bars) and mean output tokens (lines) across
 difficulty under DICoTA standard on MMAU and MMAR. \MH{it should be more clear in the legend that what bar and polyline stand for respectively.}}
  \label{fig:difficulty-aware}
\end{figure}
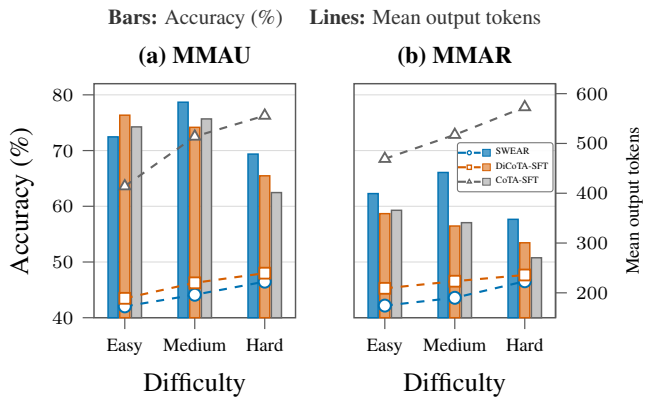

\vspace{-4pt}

Figure~\ref{fig:difficulty-aware} exhibits the intended difficulty-aware allocation behavior. As difficulty increases, our full model raises its average output length from 172.8 to 222.5 tokens on MMAU and from 174.4 to 223.2 on MMAR, while remaining 58.3--63.3\% shorter than CoTA-SFT and 5.4--16.6\% shorter than DiCoTA-SFT across all splits. With this smaller budget, it maintains competitive accuracy on easy MMAU examples and improves hard accuracy by 5.6 points over CoTA-SFT. On MMAR, it consistently outperforms both SFT variants across all difficulty levels. This trend verifies that SWEAR does not apply a uniform reasoning budget, but dynamically adjusts generation length according to estimated difficulty.

\vspace{-1em}

\subsection{Ablation Studies}

\vspace{-0.5em}

\begin{table}[ht]
  \centering
  \footnotesize
  \setlength{\tabcolsep}{3.2pt}
  \renewcommand{\arraystretch}{0.98}

  \begin{tabular*}{0.8\columnwidth}{
    @{}l
    @{\extracolsep{\fill}}
    crrrr
    @{}
  }
    \toprule
    & &
    \multicolumn{2}{c}{MMAU} &
    \multicolumn{2}{c}{MMAR} \\
    \cmidrule(lr){3-4}
    \cmidrule(l){5-6}

    Setting
    & $K$
    & Acc.$\uparrow$
    & Token$\downarrow$
    & Acc.$\uparrow$
    & Token$\downarrow$ \\
    \midrule

    \textbf{SWEAR}
    & 4
    & \second{73.5}
    & 197.1
    & \best{62.0}
    & 195.9 \\

    \addlinespace[1pt]
    \multicolumn{6}{@{}l}{\itshape Component ablations} \\
    \addlinespace[1pt]

    \quad w/o rating calibration
    & 4
    & 70.3
     & \second{188.5}
    & 59.4
    & 193.6 \\

    \quad w/o length shaping
    & 4
    & \best{74.2}
    & 216.6
    & 60.4
    & 225.5 \\


    \quad w/o adaptive KL
    & 4
    & 71.6
    & 198.4
    & \second{61.5}
    & 197.7 \\

    \addlinespace[1pt]
    \multicolumn{6}{@{}l}{\itshape Number of curriculum buckets} \\
    \addlinespace[1pt]

    \quad uniform training
    & 1
    & 69.3
    & 199.0
    & 58.2
    & \second{188.2} \\

    \quad 3 buckets
    & 3
    & 70.1
    & 198.7
    & 60.5
    & 191.4 \\

    \quad 5 buckets
    & 5
    & 71.2
    & 196.0
    & 60.8
    & 195.8 \\

    \quad 10 buckets
    & 10
    & 70.4
    & \best{180.2}
    & 59.3
    & \best{185.7} \\

    \bottomrule
  \end{tabular*}
    \vspace{-1em}
  \caption{
    Curriculum RL component ablations and sensitivity to the number of curriculum buckets.
  }
  \label{tab:ablation}
\end{table}

We further invstigate the impact of different designs and curriculum buckets. As shown in Table~\ref{tab:ablation}, removing rating calibration reduces accuracy by 3.2 and 2.6 points on MMAU and MMAR. Removing length shaping increases output length by 19.5 and 29.6 tokens, improving MMAU accuracy by 0.7 points but reducing MMAR accuracy by 1.6 points. Static KL also lowers accuracy on both benchmarks. Among curriculum settings, four buckets achieve the highest accuracy. These ablations confirm that rating prediction, adaptive length control, and curriculum scheduling each contribute to the final performance, validating the design of the proposed training strategy.

\vspace{-0.5em}

\section{Conclusion}

We proposed SWEAR, a structured difficulty modeling framework for adaptive audio reasoning. Building on CoTA, we augment each trace with multi-dimensional difficulty annotations and distill them into the LALM's self-ratings during SFT. These ratings subsequently guide curriculum RL, adapting both reasoning length and sectional demands. Experiments on MMAU and MMAR show that SWEAR produces concise traces for easy problems while preserving sufficient deliberation for hard ones. Extending this difficulty-aware principle to more diverse datasets and multimodal reasoning domains is our future direction.

\section{Compliance with Ethical Standards}

This is a numerical simulation study for which no ethical approval was required.
\bibliographystyle{IEEEbib}
\bibliography{strings,refs}

\end{document}